\documentclass[lettersize,journal]{IEEEtran}
\usepackage{amsmath,amsfonts}
\usepackage{algorithm}
\usepackage{algpseudocode}
\usepackage{array}
\usepackage[caption=false,font=normalsize,labelfont=sf,textfont=sf]{subfig}
\usepackage{textcomp}
\usepackage{stfloats}
\usepackage{url}
\usepackage{verbatim}
\usepackage{listings}
\usepackage{graphicx}
\usepackage{cite}
\usepackage{amsmath}
\usepackage{booktabs}
\usepackage{bm}
\usepackage{subfig}
\usepackage{multirow}
\usepackage{xcolor}  
\usepackage{pifont}
\usepackage[pagebackref=true,breaklinks=true,colorlinks,bookmarks=false]{hyperref}
\begin{document}

\title{Implicit and Explicit Language  Guidance for Diffusion-based Visual Perception}

\author{Hefeng Wang, Jiale Cao, Jin Xie, Aiping Yang, and Yanwei Pang
\thanks{H. Wang, J. Cao, A. Yang, and Y. Pang are with the School of Electrical and Information Engineering, Tianjin University, Tianjin 300072, China, also with Shanghai Artificial Intelligence Laboratory, Shanghai 200232, China  (E-mail: \{wanghefeng, connor, yangaiping, pyw\}@tju.edu.cn).}
\thanks{J. Xie is with the School of Big Data and Software Engineering, Chongqing University, Chongqing 401331, China (E-mail: xiejin@cqu.edu.cn).}}

\markboth{Journal of \LaTeX\ Class Files,~Vol.~14, No.~8, August~2021}%
{Shell \MakeLowercase{\textit{et al.}}: A Sample Article Using IEEEtran.cls for IEEE Journals}

\maketitle

\begin{abstract}
Text-to-image diffusion models have shown powerful ability on   conditional image synthesis. With large-scale vision-language pre-training, diffusion models are able to generate high-quality images with rich texture and reasonable structure under different text prompts. However, it is  an open problem to adapt the pre-trained  diffusion model for visual perception. In this paper, we propose an implicit and explicit language guidance framework for diffusion-based perception, named IEDP. Our  IEDP comprises an implicit language guidance branch and an explicit language guidance branch. The implicit branch employs frozen CLIP image encoder to directly generate implicit text embeddings that are fed to diffusion model, without using explicit text prompts. The explicit branch utilizes the ground-truth  labels of corresponding images as text prompts to condition feature extraction of diffusion model. During training, we jointly train diffusion model by sharing the model weights of these two branches. As a result,  implicit and explicit branches can jointly guide feature learning. During inference, we only employ implicit  branch for final prediction, which does not require any ground-truth labels. Experiments are performed on two typical perception tasks, including semantic segmentation and depth estimation. Our IEDP achieves promising performance on both tasks. For semantic segmentation, our IEDP has the mIoU$^\text{ss}$ score of 55.9\% on AD20K validation set, which outperforms the baseline method VPD by 2.2\%. For depth estimation, our IEDP outperforms the baseline method VPD with a relative gain of 11.0\%.

\end{abstract}

\begin{IEEEkeywords}
Diffusion model, language guidance, visual perception
\end{IEEEkeywords}

\section{Introduction}
\IEEEPARstart{D}iffusion models have exhibited exceptional performance in the related field of image synthesis \cite{cao2024DiffFashion,Liu2024CVDN,saharia2022photorealistic,yu2022scaling,ramesh2022hierarchical}, which can recover high-quality images from noisy data. To reduce the significant  training cost in  pixel space, stable diffusion model \cite{rombach2022high} has been proposed to perform diffusion forward and reverse backward on feature latent space. Through training on large-scale image-text paired dataset LAION-5B\cite{schuhmann2022laion}, stable diffusion model  can generate various high-quality images with rich texture and reasonable structure under various text prompts. This phenomenon suggests that stable diffusion model has the strong feature representation ability of both low-level and high-level semantic information.

\begin{figure}[t]
\centering
\includegraphics[width=0.99\linewidth]{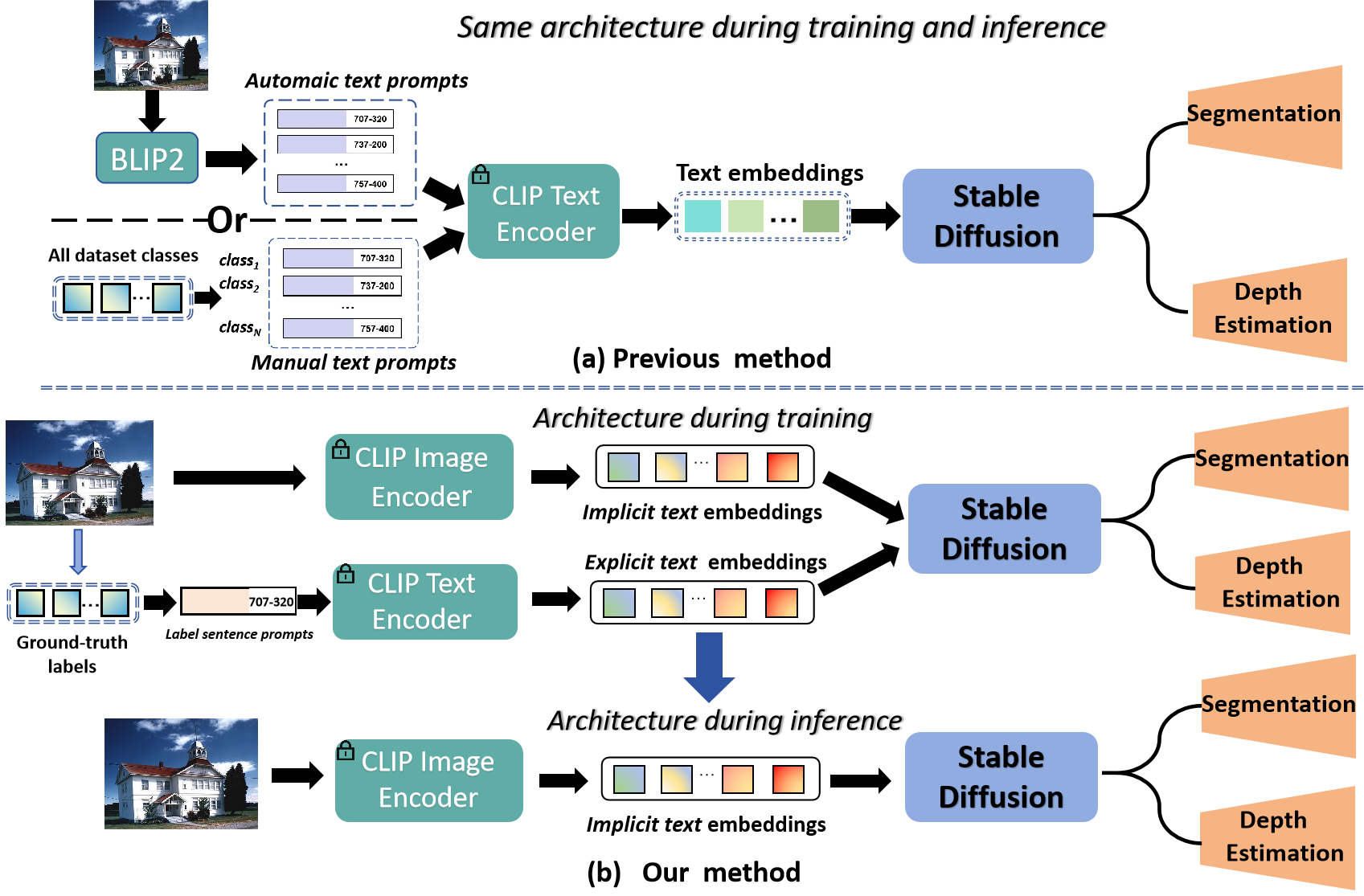}
\caption{\textbf{Comparison between existing methods and our proposed method.} In (a), the existing methods \cite{zhao2023unleashing,kondapaneni2023textimage} first employ all dataset classes or BLIP-2\cite{li2023blip} model to manually or automatically generate text prompts, and then utilize frozen CLIP\cite{radford2021learning} text encoder to extract text embeddings, which are fed to stable diffusion to condition feature extraction. During training and inference, these methods adopt the same structure. In (b), our proposed method introduces two branches to generate implicit and explicit text embeddings for stable diffusion during training, where these two branches can jointly train the model. During inference,  we only employ implicit branch  to generate implicit text embeddings for perception tasks.}
\label{fig:diffframe}
\end{figure}

Recently, the researchers \cite{zhao2023unleashing,kondapaneni2023textimage} started to explore diffusion model for visual perception via fully utilizing its  strong feature representation ability. VPD \cite{zhao2023unleashing} first generates  text prompts with all semantic classes existing in the dataset, and then utilizes these text prompts to condition feature extraction of stable diffusion. VPD has exhibited promising performance on various perception tasks, especially when using fewer training iterations. However, VPD employs all semantic classes from the entire dataset, which does not align the specific image that only contains several classes. To address this issue, TADP \cite{kondapaneni2023textimage} adopts the pre-trained vision-language model BLIP-2 \cite{li2023blip} to generate image-aligned captions, and employs these image-aligned captions to condition feature extraction of diffusion model. Compared to VPD, TADP avoids the interference of non-existing classes when conditioning feature extraction in some degree. We summarize the pipelines of VPD and TADP into Fig. \ref{fig:diffframe}(a). They initially rely on manual design  or automatic generation to create  text prompts. Subsequently, they utilize a CLIP text encoder to extract text embeddings, and fed them into stable diffusion to condition feature extraction. We argue that there exists some limitations in these methods. First, during inference, it is relatively cumbersome in TADP to employ two models to generate text prompts and text embeddings, and there exists a misalignment issue between text prompt and image in VPD. Second, during training, these methods do not fully utilize ground-truth  labels corresponding to specific training images, which accurately align with the training images.

To address these  issues above, we propose an implicit and explicit language guidance framework for diffusion-based perception, named IEDP.  Our proposed IEDP contains two different branches: implicit language guidance branch and explicit language guidance branch. In the implicit branch, we introduce a novel implicit prompt module that directly generates implicit text embeddings using CLIP image encoder, instead of  using explicit text prompts. The implicit text embeddings are fed to stable diffusion to condition feature extraction. Because CLIP model is able to learn connection between image-text pairs, the implicit text embeddings contains class  information existing in the image. In the explicit branch, we introduce an explicit prompt module to directly employ ground-truth  labels of corresponding training image as explicit text prompts. The ground-truth labels provide accurate class information of the image, which can better guide feature learning. As shown in Fig. \ref{fig:diffframe}(b), during training, we jointly train the network via weight sharing between these two branches. During inference, we only employ implicit branch for perception prediction, and remove the  explicit branch since the accurate ground-truth labels of test image are not available. We perform the experiments on typical visual perception tasks, including semantic segmentation and depth estimation. The experimental results demonstrate the efficacy of our proposed method. The contributions and merits can be summarized as follows.

\begin{itemize}
    \item We propose a novel implicit and explicit language guidance framework for diffusion-based  perception. The implicit and explicit branches jointly train the model via sharing model weights during training, while only implicit branch is employed during inference.
    \item In implicit branch, we directly employ frozen image encoder to generate implicit text embeddings to condition feature extraction, which does not require additional model to create explicit text prompts.
    \item In explicit branch, we  employ ground-truth  labels of corresponding training image to create text prompts. Compared to automatic text prompts, ground-truth based text prompts are noise-free and can better guide feature learning during training. 
    \item Our  method achieves superior performance on semantic segmentation and depth estimation. Our IEDP achieves mIoU$^\text{ss}$ score of 55.9\% on ADE20K\cite{Zhou_ADE20K_CVPR_2017} set,  outperforming diffusion-based  method VPD \cite{zhao2023unleashing} by 2.2\%. On depth estimation, our IEDP achieves  RMSE of 0.226  on NYUv2\cite{silberman2012indoor},  outperforming VPD  by 11.0\% relatively.  In addition, compared to VPD and TADP, our IEDP has the better trade-off between performance and inference time.
 
\end{itemize}

\section{Related Works}
\label{sec:relatedworks}
In this section, we first introduce the related works of visual perception. Afterwards, we give a review on diffusion models. Finally, we introduce diffusion-based visual perception.

\subsection{Visual Perception}
As a typical perception task, semantic segmentation \cite{lu2024prompt,ma2024tfrnet} aims to group the pixels in an image into different semantic categories. In past years, semantic segmentation has achieved remarkable progress, which can be mainly divided into CNN-based and transformer-based approaches. Most CNN-based approaches were developed on fully convolutional framework \cite{Long_FCN_CVPR_2015}. To improve segmentation performance, some researchers explored to exploit local contextual information using encoder-decoder  \cite{Zhao_PSPNet_CVPR_2017,Cao_TripleNet_CVPR_2019,Wang_HRNet_TPAMI_2019} or spatial pyramid structure \cite{Chen_DeepLabV3_arXiv_2017,deeplabV2}, while some other researchers proposed to exploit non-local contextual information using attention mechanism \cite{Fu_DANet_CVPR_2019,Huang_CCNet_ICCV_2019,Yuan_OCNet_IJCV_2021}. Recently, with the success of transformer in image classification \cite{dosovitskiy2020vit}, the researchers started to explore transformer-based approaches. Some approaches replace the convolutional backbone with the transformer backbone \cite{Yuan_HRFormer_NeurIPS_2021,Gu_HRViT_CVPR_2022,Xie_SegFormer_NeurIPS_2021}, while some other approaches design query-based transformer decoder for semantic segmentation \cite{Zheng_SETR_CVPR_2021,Strudel_Segmeter_ICCV_2021}.

Monocular depth estimation is another fundamental perception task, which plays an important role in 3D vision task. Eigen \textit{et al.} \cite{eigen2014depth} first proposed to employ multi-scale deep features for depth estimation. Afterwards, the researchers proposed many variants for improved depth estimation, such as designing novel model structure \cite{Xu_2018_CVPR,Bhat2021cvpr} and considering geometric constraints \cite{long2020occlusion,Yin2019enforcing,Yin2021cvpr}.

\subsection{Diffusion Models}
 
Diffusion models \cite{ho2020denoising,liu2022pseudo,lu2022dpm,Song_DDIM_ILCR_2021} have exhibited exceptional synthesis quality and controllability in the field of image synthesis. The diffusion model contains a forward diffusion process and a reverse denoising process. The forward diffusion process progressively adds the noise to the image, and the reverse denoising process tends to recover the clean image from the noise. At first, these diffusion models directly operate on pixel space, which require significant training resources and long training time. To address this issue, Rombach \textit{et al.} \cite{rombach2022high} propose the latent diffusion model (LDM) that performs diffusion on latent feature space, named stable diffusion. Stable diffusion first employs a pre-trained auto-encoder to extract latent feature representation, and then performs diffusion and  denoising processes on latent space. Finally, stable diffusion predicts the image from latent space using a decoder. In addition, stable diffusion can condition image synthesis based on texts and semantic  maps. Stable diffusion has achieved great success on conditional image synthesis.

\subsection{Diffusion-based Visual Perception}
With the success of diffusion model on image synthesis, the researchers started to employ diffusion models for visual perception. Some researchers explored to extend the diffusion pipeline for visual perception. Chen \textit{et al.} \cite{chen2023generalist} proposed to treat panoptic segmentation as a discrete data generation and employ bit diffusion to predict discrete data. Chen \textit{et al.} \cite{Chen_DiffusionDet_arXiv_2022} introduced diffusion model for object detection by  predicting the bounding-boxes from noisy bounding-boxes. Amit \textit{et al.} \cite{Amit_SegDiff_arXiv_2021} first extracted deep features from input image and then summed the deep features with noisy segmentation map to condition the denoising process. Ji \textit{et al.} \cite{ji2023ddp} proposed to concatenate the noisy image and deep features to condition the denoising process. In addition, some researchers \cite{baranchuk2021label} employed diffusion model for label-efficient segmentation.

Instead of using diffusion pipeline, some researchers \cite{zhao2023unleashing,kondapaneni2023textimage} explored to make full use of the strong feature representation ability of pre-trained diffusion model. These researchers treated diffusion model as a feature extractor. Zhao \textit{et al.} \cite{zhao2023unleashing} explored to use all dataset classes as text prompts to condition feature extraction of diffusion model, and add a task-specific decoder for visual prediction. Kondapaneni \textit{et al.} \cite{kondapaneni2023textimage} argued that using all dataset classes as text prompt can not align the specific image very well. Instead of all dataset classes, Kondapaneni \textit{et al.} \cite{kondapaneni2023textimage} proposed to employ BLIP-2 to generate image captions, and condition feature extraction using the generated image captions. However, the proposed method TADP relies on additional network to generate text descriptions during inference. In open-vocabulary panoptic segmentation, Xu \textit{et al.} \cite{Xu2023ODISE} employed CLIP image encoder to generate text embeddings, named ODISE. Compared to ODISE, our proposed method has following differences: (1) We introduce a novel implicit and explicit language guidance framework, which employs implicit and explicit prompt modules to jointly train the diffusion model for improved visual perception. (2) Compared to ODISE that employs a linear layer to generate text embeddings, our proposed method explores implicit module with  learnable queries to extract text embeddings.  (3) Instead of open-vocabulary  segmentation, our proposed method focuses on classical visual perception tasks.

\begin{figure*}[t]
\centering
\includegraphics[width=1.0\linewidth]{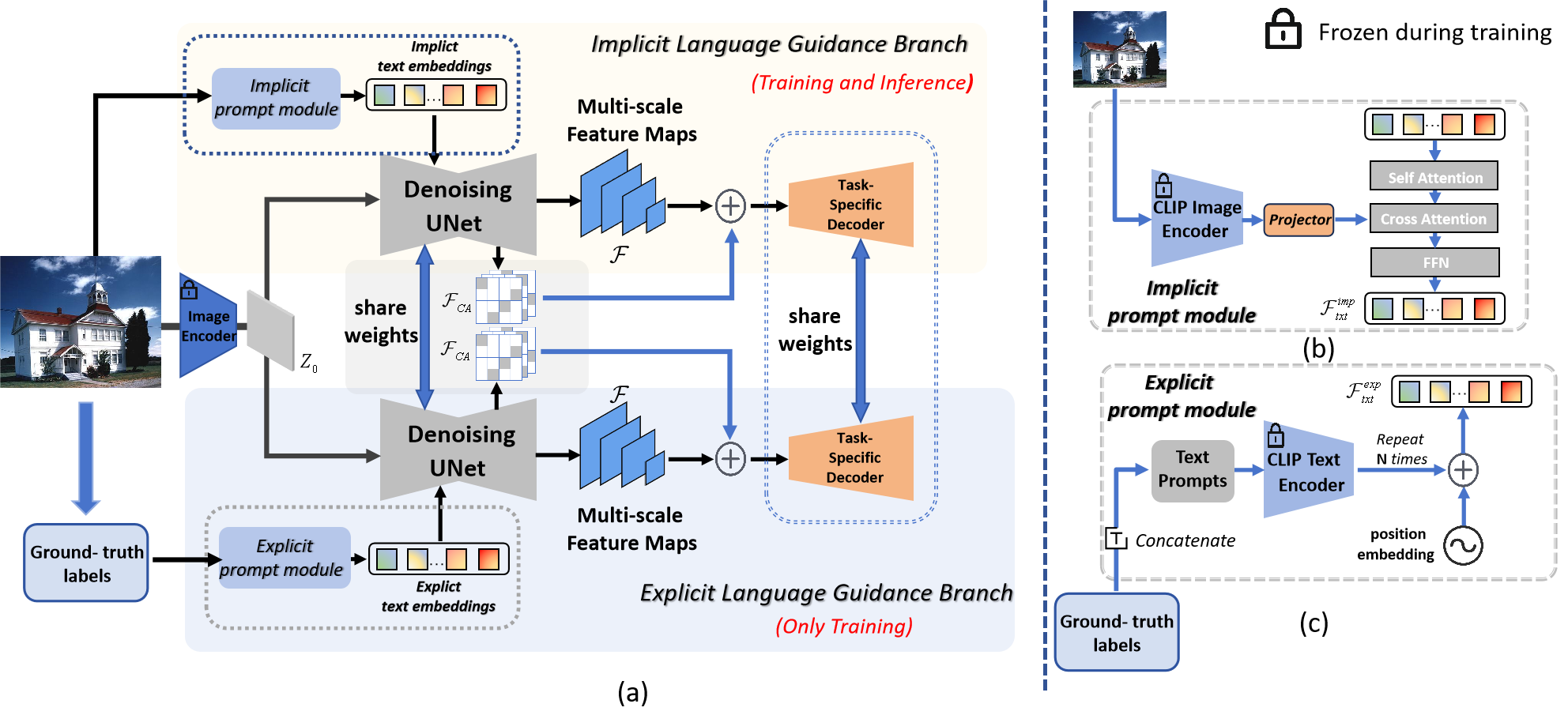}
\caption{\textbf{Overall architecture of our proposed method}. In (a), we present the overall architecture of our method. We introduce an implicit language guidance branch and an explicit language guidance branch, which respectively utilize the implicit prompt module and explicit prompt module to condition feature extraction of denoising UNet for the following task-specific decoder. These two branches share the weights of model parameters during training. During inference, we only employ the implicit branch. In (b) and (c), we give the detailed structures of implicit prompt module and explicit prompt module.}
\label{fig:frame}
\end{figure*}
\section{Method}
\label{sec:ours}

In this section, we introduce our proposed implicit and explicit language guidance framework for diffusion-based visual perception, named IEDP. Following VPD \cite{zhao2023unleashing}, our IEDP employs text-to-image stable diffusion model  
 \cite{rombach2022high} to fully utilize its strong feature representation ability. VPD employs all classes existing in dataset as text prompts, resulting in the misalignment with the specific image due to including non-existing classes. Instead of using unaligned text prompts like VPD, our IEDP introduces an explicit  prompt module and an implicit  prompt module, which both incorporate image-specific semantic information to condition feature extraction. Specifically, the explicit  prompt module employs ground-truth  labels of corresponding images to extract explicit text embeddings to condition  feature extraction of diffusion model. The implicit  prompt strategy first employs frozen CLIP image encoder to extract visual features, and then converts these visual features to implicit text embeddings as the input to stable diffusion model. 

\noindent\textbf{Overall architecture.} Fig. \ref{fig:frame} presents the overall architecture of our proposed IEDP, integrating our explicit  and  implicit  prompt modules into the base model that contains an image encoder, denoising UNet, and task-specific decoder. 
Therefore, our IEDP contains two branches: implicit language guidance branch and explicit language guidance branch, which share the weights of model parameters. Given an input image $I$, we first employ the image encoder to extract the latent features $\bm{z}_0$, which are then fed to the denoising UNet in both explicit and implicit language guidance branches. In implicit language guidance branch, we employ the implicit prompt module to extract implicit text embeddings $\mathcal{F}_{txt}^{imp}$ from the input image. Based on implicit text embeddings $\mathcal{F}_{txt}^{imp}$, we condition denoising UNet to generate hierarchical feature maps \{$\mathcal{F}_{i},i=1,2,3,4$\} and averaged cross-attention feature maps $\mathcal{F}_{CA}$ similar to VPD. Afterwards, we employ a task-specific decoder for visual  prediction. In explicit language guidance  branch, we directly unitize the explicit prompt module to generate explicit text embeddings $\mathcal{F}_{txt}^{exp}$ from the ground-truth (GT) labels of input image, and condition denoising UNet to generate hierarchical feature maps and averaged cross-attention feature maps. Similar to implicit branch, we perform the prediction with task-specific decoder.

We jointly train the model using both the implicit language guidance branch and explicit language guidance branch. During inference, we only employ implicit branch for prediction, because the ground-truth  labels of image are not available, and it is relatively time-consuming to employ two branches for prediction.

\subsection{Implicit Language Guidance Branch} 
\label{sec:ilg}

VPD \cite{zhao2023unleashing} adopts the unaligned text prompts generated by all dataset classes. Instead of using unaligned text prompts, TADP \cite{kondapaneni2023textimage} generates image-aligned text prompts using BLIP-2 generated captions. Based on automatically generated text prompts, TADP further employs a frozen CLIP text encoder to generate text embeddings fed to denoising UNet in stable diffusion model. Though TADP can adaptively generate image-aligned text prompts, we argue that it is relatively cumbersome to employ two networks to generate text prompts and text embeddings respectively. To address this issue, we propose an implicit prompt module to directly generate image-aligned text embeddings from the input image  using a single  network.

Fig. \ref{fig:frame}(b) shows the structure of our implicit prompt module. Specifically, we  employ frozen CLIP image encoder  and a light-weight projector to extract the image visual features $\mathcal{F}_{vis}$, which can be written as
\begin{equation}
\mathcal{F}_{vis} =f_{pro}(f_{CLIP}^{img}(I)),
\end{equation}
where  the projector $f_{pro}$ is an MLP layer. Because CLIP is able to learn  the connection between image-text pairs, we believe that the visual features $\mathcal{F}_{vis}$ implicitly contain semantic information to learn text embeddings.
Therefore, we employ a set of learnable queries $\mathcal{Q}$ to predict implicit text embeddings from the visual features $\mathcal{F}_{vis}$. Specifically, we first feed the learnable queries $\mathcal{Q}$ to a self-attention layer. Then, we perform a cross-attention operation between learnable queries and visual features. Finally, we use an MLP layer to generate implicit text embeddings $\mathcal{F}_{txt}^{imp}$. We summarize the detailed process as 
\begin{equation}
\begin{aligned}
&\mathcal{Q}_{s}=\mathcal{Q}+f_{SA}(f_{LN}(\mathcal{Q})),\\
&\mathcal{Q}_{c}=\mathcal{Q}_{s}+f_{CA}(f_{LN}(\mathcal{Q}_{s}),f_{LN}(\mathcal{F}_{vis})),\\
&\mathcal{F}_{txt}^{imp}=\mathcal{Q}_{c}+f_{FFN}(f_{LN}(\mathcal{Q}_{c})),
\end{aligned}
\end{equation}
where $f_{SA}$ is self-attention operation, $f_{CA}$ is cross-attention operation, $f_{LN}$ is layer norm operation, and $f_{FFN}$ is feed-forward layer.  We employ $\mathcal{F}_{txt}^{imp}$ as implicit text embeddings because it is generated from CLIP image encoder, instead of text encoder.

Based on the implicit text embeddings $\mathcal{F}_{txt}^{imp}$ and the latent features $\bm{z}_0$, we employ denoising UNet  to generate hierarchical feature maps and averaged cross-attention map. Similar to VPD, we concatenate these features and feed them to task-specific decoder head for prediction.

\subsection{Explicit Language Guidance Branch}
Compared to unaligned text prompts, the image-aligned text prompts can better condition  feature extraction of denoising UNet for visual perception tasks. Though TADP \cite{kondapaneni2023textimage} employs BLIP-2 to generate image-aligned text prompts for each image, it is still not very accurate and has noise. In fact, the training image has the corresponding ground-truth class labels, which can be used to generate more accurate text prompts. However, the ground-truth labels of test image are not available. Therefore, it is a natural question, can we only use the accurate ground-truth labels of training images to help improve visual perception model learning during training. Inspired by this, we introduce an explicit prompt module  to jointly guide feature learning with implicit branch.

As shown in Fig. \ref{fig:frame}(c),  we first concatenate all ground-truth  labels in an image as  text prompts $\mathcal{P}$, which can be written as 
\begin{equation}
\mathcal{P}=\{c + ``~" | c\in \mathcal{C}_{gt} \},
\end{equation}
where $c$ represents the word of class label, and $\mathcal{C}_{gt}$ represents the set of all class labels in an image. Then, we feed the  text prompts $\mathcal{P}$ to the frozen CLIP text encoder to generate explicit text embeddings $\mathcal{F}_{txt}^{exp}$ as
\begin{equation}
\mathcal{F}_{txt}^{exp} =f_{CLIP}^{txt}(\mathcal{P}).
\end{equation}
To keep the number of explicit text embeddings be consistent with the implicit  text embeddings, we repeat the explicit text embeddings. Afterwards, we employ the denoising UNet and task-specific decoder for prediction.

Compared to image-aligned text prompts in TADP, our text prompts are clean and noise-free, which can better guide feature learning.  By sharing model weights with implicit branch, the explicit  branch is able to jointly affect the feature learning of denoising UNet and task-specific decoder. Moreover, different to TADP, we only employ explicit branch during training, and remove the explicit branch during inference.

\subsection{Training and Inference }
During training, we set the time step $t$ to be zero, which results in the latent feature map $\bm{z}_{0}$ being noise-free. The hierarchical features $\mathcal{F}$ can be obtained by extracting the last layer of each decoder block in denoising U-Net, consisting of 4 feature maps of different resolutions. We represent these four feature maps as $\mathcal{F}_1,\mathcal{F}_2,\mathcal{F}_3,\mathcal{F}_4$, which has the stride of 8, 16, 32, 64 to the input image. The averaged cross-attention map $F_{CA}$ are averaged by the cross-attention maps in both encoder and decoder blocks. Afterwards, the hierarchical feature maps and averaged cross-attention map are fed to task-specific decoder to generate the outputs. Finally, we calculate the overall training loss $L$, including the implicit loss $L_{imp}$ and explicit loss  $L_{exp}$ in implicit and explicit branches, which can be written as
\begin{equation}
L= L_{imp}+ L_{exp}.
\end{equation}
The specific implicit loss $L_{imp}$ and explicit loss $L_{exp}$ are different for different perception tasks. We use the cross-entropy loss for semantic segmentation task, and the
Scale-Invariant loss (SI) \cite{eigen2014depth}   for depth estimation.

Our proposed method does not rely on any additional text information during inference, setting it apart from previous methods that use manual text templates or automatic image captions. We directly feed the image  to implicit language guidance branch for prediction. Specifically, we  use CLIP image encoder to generate implicit text embeddings, and  stable diffusion model to generate hierarchical features and cross-attention features, which are fed to a task-specific decoder to generate the final prediction.

\begin{table*}[t!]
\centering
\footnotesize
\caption{\textbf{Semantic segmentation comparison with some other methods on  ADE20K.} The compared methods include non-diffusion-based (classical) approaches and diffusion-based approaches. Compared to these diffusion-based approaches, our method  achieves a superior performance.}
\begin{tabular}{r|c|c|c|cc}
\toprule
Method&Publication & Crop  & mIoU$^\text{ss}\uparrow$ & mIoU$^\text{ms}\uparrow$  \\
\midrule
\multicolumn{4}{l}{\textit{Non-diffusion-based approaches:\hfill}} \\
Swin-L \cite{Liu_Swin_ICCV_2021}&ICCV’21   &   $640^{2}$ & 52.1 & 53.5  \\
ConvNeXt-L \cite{liu2022convnet}&CVPR’22     &  $640^{2}$ & 53.2 & 53.7  \\
ConvNeXt-XL~\cite{liu2022convnet}&CVPR’22   &  $640^{2}$ & 53.6 & 54.0   \\

CLIP-ViT-B~\cite{rao2022denseclip}&CVPR’22  &  $640^{2}$ & 50.6 & 51.3  \\
ViT-Adapter-L\cite{chen2022vision}&ICLR’23   &  $640^{2}$ & 56.8 & 57.7  \\
\midrule
\multicolumn{4}{l}{\textit{Diffusion-based approaches:\hfill}} \\
DDP\cite{ji2023ddp} &ICCV’23&$512^{2}$&53.2&54.4\\
VPD\cite{zhao2023unleashing} &ICCV’23  &  $512^{2}$& 53.7 & 54.6  \\
TADP\cite{kondapaneni2023textimage}& CVPR’24 &   $512^{2}$& 54.8 & 55.9  \\
\textbf{IEDP (Ours)} & - &   $512^{2}$ &  55.9 & 57.1   \\

\bottomrule
\end{tabular}
\label{tab:semantic}
\end{table*}

\section{Experiments}
\label{sec:exp}
In this section, we perform experiments to demonstrate the efficacy and superiority of our proposed method. 
We evaluate our method on two typical perception tasks, including semantic segmentation and depth estimation.
\subsection{Datasets and Evaluation Metrics}

\noindent\textbf{ADE20K.}  The ADE20K dataset \cite{Zhou_ADE20K_CVPR_2017} is a classical  semantic segmentation dataset collected from the website. It contains more than 20$K$ natural images with pixel-level annotations. There are totally 150 semantic categories. There are three subsets, including $\texttt{train}$, $\texttt{val}$, and $\texttt{test-dev}$ sets. The training set has 20210 images, $\texttt{val}$ set has about 2000 images, and the $\texttt{test}$ set has about 3000 images. Similar to VPD, we train our method on $\texttt{train}$ set and compare with other methods on $\texttt{val}$ set. We adopt mIoU as evaluation metric of semantic segmentation.

\noindent\textbf{NYUv2.} The NYUv2 dataset \cite{silberman2012indoor} is a typical depth estimation dataset collected from a variety of indoor scenes. The images and corresponding depth maps are captured by the pairs of RGB and Depth cameras. There are 24$k$ images for training, and 654 images for testing. We adopt three commonly used indicators for depth estimation evaluation, including  absolute relative error (REL),
root mean squared error (RMSE), and average $log10$ error between predicted depth $\hat{d}$ and ground-truth depth $d$. We further present threshold accuracy ${\delta}_{n}$ which represents
 $\delta_{n}$=\% of pixels satisfying $\max(d_{i}/\hat{d}_{i},\hat{d}_{i}/d_{i})
  <1.25^{n}$ for $n = 1,2,3$.

\subsection{Experiment Setups}

During training, we freeze the image encoder, CLIP image encoder, and CLIP text encoder. Namely, we only fine-tune the denoising UNet and task-specific decoder on four NVIDIA RTX A6000 GPUs. We provide the detailed experiment setups of different tasks as follows.

\noindent\textbf{Semantic segmentation.} We adopt the semantic FPN as decoder for semantic segmentation as in VPD.  During training, we adopt AdamW as the optimizer, and 
set the batch size as 8. We employ the poly learning  rate schedule with a power of 0.9. The initial learning rate is set as 0.00008, and the weight decay is set as 0.001. When comparing with other methods, the number of iterations are 160K. For fast ablation study, the number of iterations are set as 40K. During inference, we perform slide window inference, where the crop window size is 512 $\times$ 512 pixels, and the slide stride is 341 $\times$ 341 pixels.

\noindent\textbf{Depth estimation.} We adopt the similar decoder as in \cite{xie2023revealing}. During training, we adopt AdamW as the optimizer, and employ the poly learning rate rate schedule. We randomly crop the images
into 480$\times$480 pixels,  and set  the batch size as 24. There are totally 25 epochs  The initial learning rate is 5e-4 and the weight decay as 0.1. Because depth estimation dataset does not provide ground-truth labels, we employ BLIP to generate image captions fed to explicit language guidance branch.
During inference, we employ the techniques of horizontal flipping and sliding windows as in VPD.

\begin{table*}[t]
\centering
\footnotesize
\caption{\textbf{Depth estimation comparison with some other methods on NYUv2.} The compared methods include non-diffusion-based (classical) approaches and diffusion-based approaches. Compared to DDP and VPD, our proposed method also has superior results. Compared to TADP, our proposed method does not require the heavy model BLIP-2 and explicit text encoder.}
\begin{tabular}{r|c|cccccc}
\toprule
Method&Publication & RMSE$\downarrow$ & ${\delta}_{1}\uparrow$ & ${\delta}_{2}\uparrow$ & ${\delta}_{3}\uparrow$ & REL$\downarrow$ & ${log}{10}$$\downarrow$ \\ 
\midrule
\multicolumn{6}{l}{\textit{Non-diffusion-based approaches:\hfill}} \\
BTS \cite{lee2019big}& ArXiv’19& 0.392 & 0.885  & 0.978 & 0.995 &  0.110 & 0.047\\
AdaBins\cite{bhat2021adabins}&CVPR’21& 0.364& 0.903& 0.984& 0.997 &0.103& 0.044\\
DPT\cite{ranftl2021vision}&ICCV’21 &0.357& 0.904& 0.988& 0.998& 0.110& 0.045\\
P3Depth\cite{patil2022p3depth}&CVPR’22 &0.356 &0.898 &0.981 &0.996 &0.104 &0.043\\
NeWCRFs \cite{yuan2022new}&CVPR’22 &0.334 &0.922 &0.992 &0.998& 0.095 &0.041\\
AiT \cite{ning2023all}&ICCV’23 &0.275 &0.954& 0.994 &0.999 &0.076 &0.033\\
ZoeDepth \cite{bhat2023zoedepth}&ArXiv’23 &0.270 &0.955 &0.995 &0.999 &0.075 &0.032\\

\midrule
\multicolumn{6}{l}{\textit{Diffusion-based approaches:\hfill}} \\
DDP\cite{ji2023ddp} &ICCV’23&0.329&0.921&0.990&0.998&0.094&0.040\\
VPD\cite{zhao2023unleashing}&ICCV’23 &0.254 &0.964 &0.995 &0.999& 0.069 &0.030\\
TADP\cite{kondapaneni2023textimage}& CVPR’24 &0.225 &0.976 &0.997 &0.999 &0.062 &0.027\\
\textbf{IEDP (Ours)}&- & 0.226  & 0.975 & 0.997  & 0.999   & 0.062  & 0.027 \\
\bottomrule
\end{tabular}
\label{tab:sota_instance}
\end{table*}

\begin{table*}[t]
\centering
\footnotesize
\caption{\textbf{Comparison in terms of both performance and inference time.} We compare three diffusion-based approaches, including VPD, TADP, and our IEDP, on both semantic segmentation and depth estimation. Because TADP does not release the source code, we report TADP inference time by our implementation that replaces manual text prompts in VPD with BLIP-2 based text prompts having the minimum length of 40. Our proposed method has the better trade-off between performance and speed.}
\begin{tabular}{r|c|cc|cc}
\toprule
\multirow{2}{*}{Method}  & \multirow{2}{*}{Publication}   &\multicolumn{2}{c}{Semantic segmentation (ADE20K)}  & \multicolumn{2}{|c}{Depth estimation (NYUv2)}\\
\cmidrule{3-6}
& &mIoU$^\text{ss}$ $\uparrow$  & Time (s) $\downarrow$ &RMSE $\downarrow$ &Time (s) $\downarrow$ \\
\midrule
VPD \cite{zhao2023unleashing} & ICCV’24 &53.7&0.274&0.254&0.261\\
TADP \cite{kondapaneni2023textimage}& CVPR’24&54.8&1.255&0.225&1.236\\
\textbf{IEDP (Ours)} & - &55.9&0.296&0.226&0.280\\
\bottomrule
\end{tabular}
\label{tab:res and spe}
\end{table*}

\subsection{State-of-the-art Comparison}
\noindent\textbf{Semantic segmentation.} We compare our proposed method with some state-of-the-art methods  on ADE20K  dataset in Table \ref{tab:semantic}. These state-of-the-art methods are divided into classical approaches and diffusion-based approaches. We report both the single-scale and multi-scale results for all these
approaches. Compared to diffusion-based approaches, our IEDP achieves significant improvement. For instance, VPD has the mIoU$^\text{ss}$ score of 53.7\%, while our IEDP has the mIoU$^\text{ss}$ score of 55.9\%. Therefore, our IEDP outperforms VPD with an absolute gain of 2.2\%  on mIoU$^\text{ss}$. Compared to TADP, our IEDP has 1.1\% improvement on mIoU$^\text{ss}$ and  1.2\% improvement on mIoU$^\text{ms}$.

\noindent\textbf{Depth estimation.}  We compare our proposed method with some state-of-the-art methods  on NYUv2  dataset in Table \ref{tab:sota_instance}. These state-of-the-art methods are also divided into classical approaches and diffusion-based approaches.  Compared to diffusion-based approaches, our IEDP achieves promising performance. For example, DDP \cite{ji2023ddp} has the RMSE score of 0.329,  VPD \cite{zhao2023unleashing} has the RMSE score of 0.254, and our IEDP has the RMSE score of 0.226. Therefore, compared to DDP and VPD, our IEDP has the relative improvements of 31.3\% and 11.0\% in terms of RMSE. Compared to TADP \cite{kondapaneni2023textimage}, our proposed IEDP has a comparable performance, but does not requires the heavy model BLIP-2 for generating text prompts and explicit text encoder.

\noindent\textbf{Inference time.}  We further compare our proposed method with diffusion-based methods in terms of both performance and inference time in Table \ref{tab:res and spe}.  We report the inference time of these methods on single NVIDIA A6000 GPU.  Compared to VPD \cite{zhao2023unleashing} and TADP \cite{kondapaneni2023textimage}, our IEDP has a better trade-off between performance and inference time. For instance, our IEDP outperforms TADP by 1.1\% while running at 4.2 times faster on semantic segmentation.

\begin{table}[t]
\centering
\footnotesize
\caption{\textbf{Impact of integrating two proposed modules into the baseline.} The baseline adopts  same architecture as VPD \cite{zhao2023unleashing}.}
\begin{tabular}{cc|cc}
\toprule
 Implicit   & Explicit  &mIoU$^\text{ss}$&mIoU$^\text{ms}$\\
\midrule
\ding{55}&\ding{55}&48.1&49.5\\
\checkmark&\ding{55}&49.8&51.1\\
\checkmark&\checkmark&52.5&53.9\\
\bottomrule
\end{tabular}
\label{tab:part_eff}
\end{table}

\begin{table}[t]
\centering
\footnotesize
\caption{\textbf{Impact of different adapter design in implicit branch.} We employ an MLP layer or learnable queries with cross-attention as the adapter.}
\begin{tabular}{c|cc}
\toprule
Design &mIoU$^\text{ss}$&mIoU$^\text{ms}$\\
\midrule
Using an MLP layer&52.1&53.2\\
Using learnable queries&52.5&53.9\\
\bottomrule
\end{tabular}
\label{tab:implict}
\end{table}

\begin{table}[t]
\centering
\footnotesize
\caption{\textbf{Impact of different numbers of learnable queries.} We set the number of learnable queries as 256.}
\begin{tabular}{c|cc}
\toprule
Number of query &mIoU$^\text{ss}$&mIoU$^\text{ms}$\\
\midrule
64&51.9&52.4\\
128&51.5&52.5\\
256&52.5&53.9\\
\bottomrule
\end{tabular}
\label{tab:query number}
\end{table}

\subsection{Ablation Study}
In this section, we will perform detailed analysis to further evaluate the effectiveness of each component in our proposed IEDP.  We conduct experiments on semantic segmentation  task using ADE20K dataset. We report the results using a fast training strategy with the training iterations of 40K.

\noindent\textbf{Effectiveness of two proposed modules.}  We first perform the ablation study to show the effectiveness of implicit and explicit prompt modules. Table \ref{tab:part_eff} presents the results of progressively integrating these two modules into the baseline. The baseline adopts the unaligned text prompts as in VPD \cite{zhao2023unleashing}, which has the mIoU$^\text{ss}$ score of 48.1\%. When replacing the unaligned text prompts with our implicit prompts, it has the mIoU$^\text{ss}$ score of 49.8\%, outperforming the baseline by 1.7\%. When further integrating our explicit prompt module, it has the mIoU$^\text{ss}$ score of 52.5\%, which outperforms the baseline by 4.4\%  totally.

\noindent\textbf{Different designs of the adapter.} Table \ref{tab:implict} shows the results of different adapter designs in implicit branch, where the adapter aims to convert the visual features of CLIP image encoder to implicit text embeddings for denoising UNet. We give two different designs, including using an MLP layer and using learnable queries to convert visual features. We observe that using learnable queries has the better performance. 

\noindent\textbf{Number of learnable queries.} Table \ref{tab:query number} shows the impact of different numbers of learnable queries in implicit branch. We observe that there exists a certain degree of performance degradation when using small number of queries, but the decline is relatively limited. Moreover, compared to the baseline, using different queries all have the improvements. We employ the number of queries as 256 as our default setting.

\begin{figure*}[t]
\centering
\includegraphics[width=\linewidth]{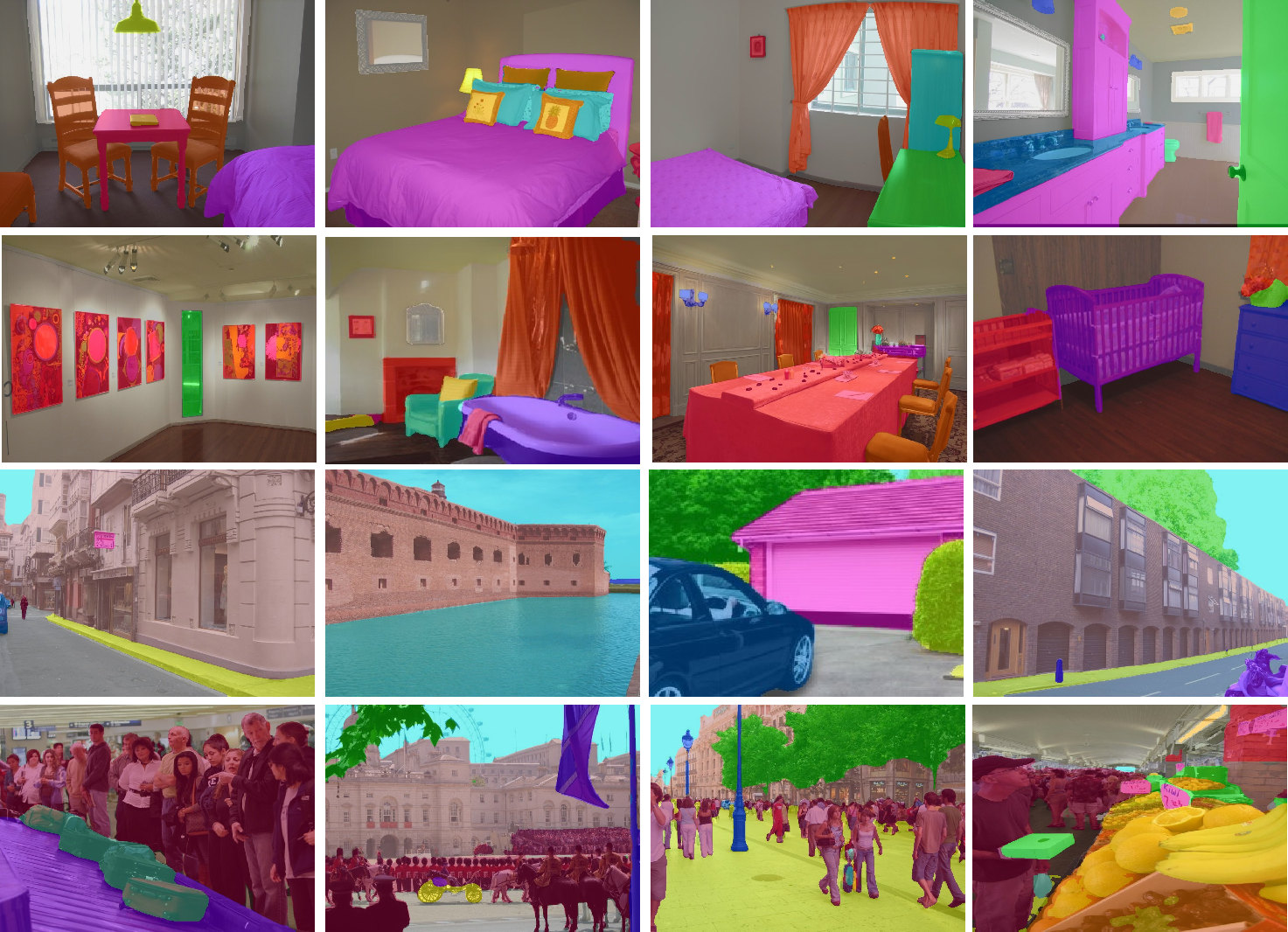}
\caption{\textbf{Visualisation results of semantic segmentation.} We provide some qualitative segmentation  examples of our proposed method on ADE20K dataset. It can be observed that our proposed method has good segmentation results in various scenarios, including indoor, outdoor, and crowded scene.}
\label{fig:semvis}
\end{figure*}

\begin{table}[t]
\centering
\footnotesize
\caption{\textbf{Impact of position embeddings.} We report the results using position embeddings or not in both implicit and explicit branches.}
\begin{tabular}{c|cc}
\toprule
Position embeddings  &mIoU$^\text{ss}$&mIoU$^\text{ms}$\\
\midrule
\ding{55}&52.3&53.8\\
\checkmark&52.5&53.9\\
\bottomrule
\end{tabular}
\label{tab:pos}
\end{table}

\begin{table}[t]
\centering
\footnotesize
\caption{\textbf{Comparison  of ground-truth  labels and BLIP-based captions.} In the explicit language guidance branch, we employ ground-truth  labels and BLIP-based captions as text prompts.}
\begin{tabular}{c|cc}
\toprule
Text prompts &mIoU$^\text{ss}$&mIoU$^\text{ms}$\\
\midrule
Using BLIP generated captions &52.0&52.6\\
Using ground-truth  labels &52.5&53.9\\
\bottomrule
\end{tabular}
\label{tab:language}
\end{table}

\noindent\textbf{Impact of position embeddings.} Table \ref{tab:pos} shows the impact of using position embeddings in both implicit and explicit branch. When not using the position embeddings, it has 0.2\% performance drop on mIoU$^{\text{ss}}$. Therefore, we add position embeddings in both two branches.

\noindent\textbf{Ground-truth labels \textit{vs.} BLIP-based captions.} In explicit language guidance branch, instead of using ground-truth labels, we can also employ BLIP\cite{li2022blip} to automatically generate image captions as text prompts like TADP. Table \ref{tab:language} compares ground-truth  labels and BLIP-based captions. We observe that it has better performance using ground-truth  labels that are less noisy.

\begin{figure*}[!t]
\centering
\includegraphics[width=\linewidth]{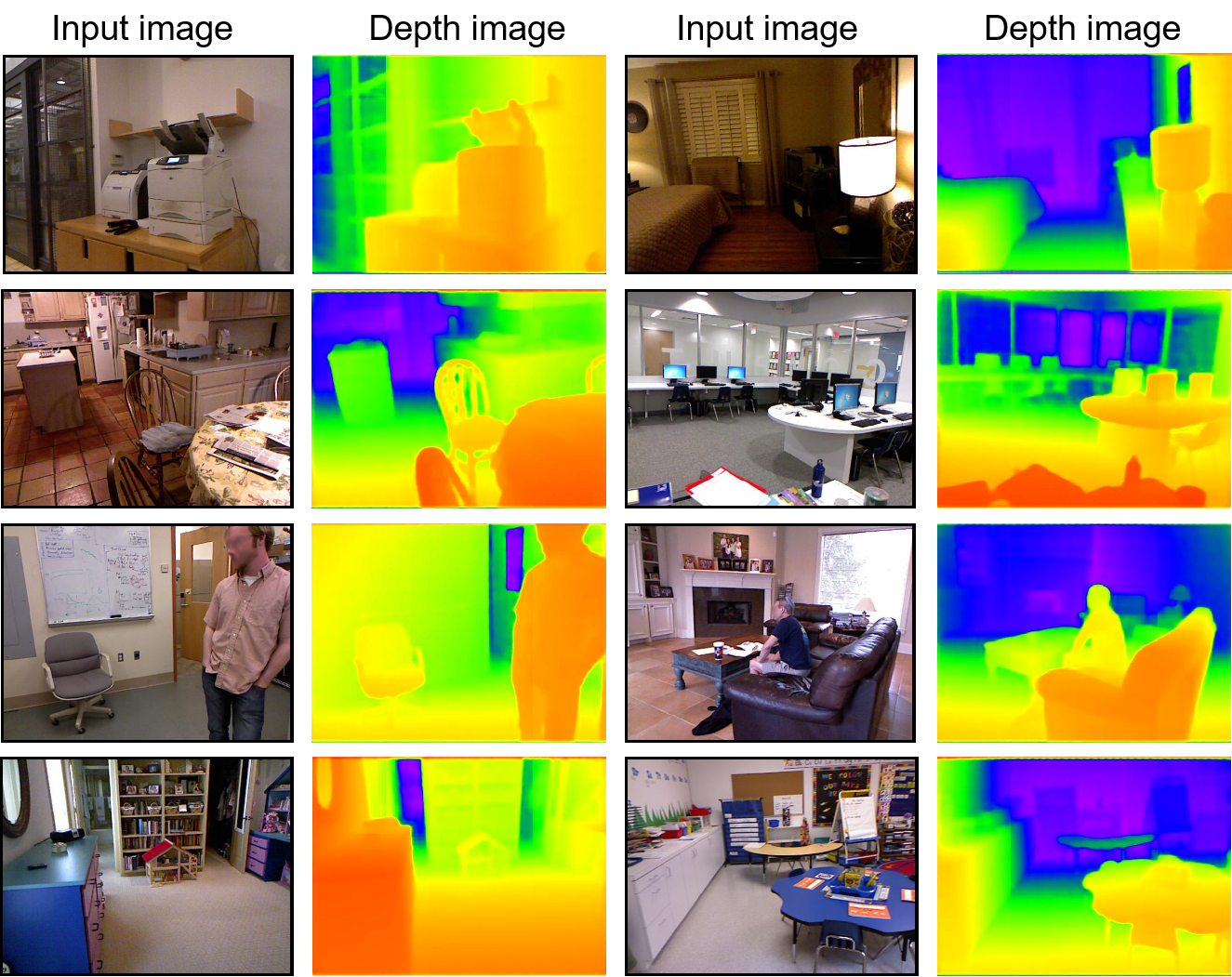}
\caption{\textbf{Visualisation results of depth  estimation.} We provide some qualitative examples of our proposed method on NYUv2 dataset, including the input images and predicted depth maps.}
\label{fig:depthvis}
\end{figure*}
\subsection{Qualitative Results}
In this subsection we present some qualitative results of our proposed method on both semantic segmentation and depth estimation tasks.

\noindent\textbf{Semantic segmentation.} Fig. \ref{fig:semvis} presents some 
semantic segmentation examples of our method on ADE20K dataset. Our proposed method presents accurate segmentation results under both indoor, outdoor, and crowded scenes.

\noindent\textbf{Depth estimation.} Fig. \ref{fig:depthvis} presents some 
depth estimation examples of our method on NYUv2 dataset. Our proposed method generates accurate depth maps for different indoor scenes, such as bedroom, office, and kitchen.

\section{Conclusion}
\label{sec:conclusion}

In this paper, we propose an implicit and explicit language guidance framework for diffusion-based visual perception, named IEDP. Our IEDP introduces an implicit prompt module and an explicit prompt module into  text-to-image diffusion model. In implicit prompt module, we employ a frozen CLIP image encoder to directly generate implicit text embeddings and fed these embeddings to diffusion model  to condition feature extraction. In explicit prompt module, we utilize the ground-truth  labels of training images as explicit text prompts and employ CLIP text encoder to generate text embeddings for diffusion model. The implicit prompt module and  explicit prompt module jointly train the diffusion model, and only implicit prompt module is used during inference. Experiments are performed on semantic segmentation and depth estimation tasks to show the efficacy of proposed method. Our proposed method achieves promising performance without using manually unaligned text prompts or automatic text prompts with heavy image caption model.

\medskip
{\small
\bibliographystyle{unsrt}
\bibliography{egbib}
}

\end{document}